\documentclass{singlecol-new}
\usepackage{natbib,stfloats}
\usepackage{mathrsfs,amsmath,upgreek}

\usepackage{natbib,stfloats}
\usepackage{mathrsfs,amsmath}
\usepackage{upgreek}
\usepackage{natbib,stfloats}
\usepackage{mathrsfs,amsmath,upgreek}
\usepackage{tabularx,multicol,longtable}
\usepackage{graphics, graphicx}

\usepackage{caption}
\usepackage{subcaption}

\newcommand{\be}{\begin{eqnarray}}
\newcommand{\ee}{\end{eqnarray}}

\theoremstyle{TH}{

}

\theoremstyle{THrm}{

}

\theoremstyle{THhit}{

}

\makeatletter

\def\theequation{\arabic{equation}}

 \makeatother

\begin{document}%
%%%%%%%%%%%%%%%%%

\thispagestyle{plain}

\setcounter{page}{1}

\LRH{xxxx}

\RRH{xxxx}

\VOL{x}

\ISSUE{x}

\PUBYEAR{xxxx}

%\BottomCatch

%\CLline

%\subtitle{}

\title{Green Heron Swarm Optimization Algorithm - State-of-the-Art of a New Nature Inspired Discrete Meta-Heuristics}

\authorA{Chiranjib Sur, Anupam Shukla}

\affA{ABV-Indian Institute of Information Technology \& Management, Gwalior\\
Email: chiranjibsur@gmail.com, dranupamshukla@gmail.com}

%\authorB{xxxx}
%\affB{xxxx}

\begin{abstract}
Many real world problems are NP-Hard problems are a very large part of them can be represented as graph based problems. This makes graph theory a very important and prevalent field of study. In this work a new bio-inspired meta-heuristics called Green Heron Swarm Optimization (GHOSA) Algorithm is being introduced which is inspired by the fishing skills of the bird. The algorithm basically suited for graph based problems like combinatorial optimization etc. However introduction of an adaptive mathematical variation operator called Location Based Neighbour Influenced Variation (LBNIV) makes it suitable for high dimensional continuous domain problems. The new algorithm is being operated on the traditional benchmark equations and the results are compared with Genetic Algorithm and Particle Swarm Optimization. The algorithm is also operated on Travelling Salesman Problem, Quadratic Assignment Problem, Knapsack Problem dataset. The procedure to operate the algorithm on the Resource Constraint Shortest Path and road network optimization is also discussed. The results clearly demarcates the GHOSA algorithm as an efficient algorithm specially considering that the number of algorithms for the discrete optimization is very low and robust and more explorative algorithm is required in this age of social networking and mostly graph based problem scenarios.
\end{abstract}

\KEYWORD{Green Heron Swarm Optimization, combinatorial optimization, graph problems, Location Based Neighbour Influenced Variation}

%\REF{to this paper should be made as follows: xxxx (xxxx) `xxxx',
%{\it xxxx}, Vol.~x, No.~x, pp.xxx--xxx.}

%\begin{bio}
%\tc{AUTHOR PLEASE SUPPLY CAREER HISTORY OF NO MORE THAN 100 WORDS
%FOR EACH AUTHOR.}
%\end{bio}

\maketitle

\section{Introduction}
\label{sec-introduction}
The present world craves for optimization with respect to every possible aspects of the nature and its happenings due to the rapid depletion of easy energy sources and profit maximization. Hence a lot of focus has been made on optimization of different fields of engineering problems and management problems mainly which are multi-dimensional and mathematical based. However there remains another sector which requires considerable attention and has represented many real life and real time problems starting from path planning, social media interaction, diffusion and ranking, recommendation systems, constraint processes scheduling, multi-objective optimization etc. This field is not a new one but has changed the way of data representation and interaction. It is better known as discrete problems (as the elements are discrete events and the conditional sequence of them held immense importance) and constitutes all problems related to graph based problems and combinatorial optimization problems. This work is marked by the introduction and application of such a discrete multi-agent based algorithm called Green Heron Swarm Optimization (GHOSA) Algorithm. It suits the discrete problems because of its combination generation  and optimization capability for the discrete problems. 

The newly introduced GHOSA is inspired by the habit of the Green Heron bird for food acquisition through their artistic skills, senses and intelligence. The algorithm is marked by the three probabilistic multi-agent based combination generation steps and an adaptive mathematical variation exclusively only for the continuous numerical mathematical problems and the various system models. The algorithm is applied on Travelling Salesman problem, 0/1 Knapsack problem and the Quadratic Assignment problems and the results are compared with the optimum values. The extended algorithm (inclusive of LBNIV) is applied on the various multi-dimensional numerical benchmark equations and optimization is sought with respect to the optimum and is compared with real coded Genetic Algorithm and Particle Swarm Optimization. Green Heron Optimization Algorithm (GHOA) %\citep{gho1}, \citep{gho2} 
denotes the algorithm without LBNIV operator and Green Heron Swarm Optimization Algorithm (GHSOA) represents the whole lot. So GHSOA signifies that the algorithm is exclusively for continuous domain problems while GHOA can be used for both discrete and limited continuous problems.

The rest of the article is arranged as Section 2 for the related work in bio-inspired computation, Section 3 with the description of the bird, Section 4 with the illustration of the GHOSA, Section 5 with details of the algorithm for the various applications and results, Section 6 for benchmark performance evaluation and scope for hybridization and Section 7 concludes with future works.

\section{Related work}
\label{sec-related}
There are several bio-inspired algorithms which are highly successful in achieving near optimal optimization of various problems of the nature both in continuous and discrete domain. Discrete Particle Swarm Optimization \citep{p1}, discrete Genetic Algorithm  \citep{p2} are specialized in generating discrete values but have limited applicability for combinatorial optimization which are handled well by Ant Colony Optimization (ACO) Algorithm \citep{p3}, \citep{acome}, Intelligent Water Drop (IWD) Algorithm \citep{p4}, Egyptian Vulture Optimization Algorithm \citep{evo1}, \citep{evo2} etc like optimization algorithms. Other continuous domain algorithms like Particle Swarm Optimization (PSO) Algorithm \citep{p5} helps in social influenced multi-agent like swarm search operations, Genetic Algorithm (GA) is specialized in combination formation out of the existing solutions and prevents local stagnancy of agents, Bat Algorithm  \citep{p6} is an enhanced multi-hierarchy based swarm intelligence, Bacteria Foraging Optimization Algorithm \citep{p8} operates on swarming like grouping movement and increases the level of local searches, Krill Herd  \citep{p9} works on krill organism like movement and search for optimization peak, Artificial Bee Colony (ABC) \citep{p10} divides the agents for cooperative exploration and exploitations and the agents switches character, Honey Bee Swarm  \citep{p11} works on reproduction and crossover of their breed for better agent generation, Firefly Algorithm  \citep{p12} works on the cooperative influence of the better agents through light attraction like feature, Glowworm Swarm Optimization  \citep{p13} also works on light attraction like feature  but the attracting agents are limited and within a range, Cuckoo Search  \citep{p14} works on random placement and removal of solution variables and thus promotes mixing and opportunity, Artificial immune system  \citep{p15} develops detection and prevention factors for optimization, Differential evolution \citep{p16} works on the principle of genetic algorithm but employs functions for combination, crossover and mutation, Differential Search \citep{p17} works on Brownian random walk movement search over the work-space, Harmony Search  \citep{p18} combination of solution variable values from a bunch of values to generate optimized solution sets, Biogeographical based optimization \citep{p19} depends on mathematical relation based migration of variable values from solution set to another to promote enhancement in solution quality, Invasive Weed Optimization \citep{p20} depends on random and deterministic variation like spreading of agent for optimization, Simulated Annealing \citep{p21} is an extension of hill-climbing algorithm which have introduced probabilistic acceptance and adaptive step size phenomenon, Honey Bee Mating Optimization  \citep{p22} utilizes the crossover feature similar to GA between bee to produce enhancement in solution quality, League Championship Algorithm  \citep{p23} employs tournament like situation to generate combinations as solutions, Teaching-Learning-Based Optimization (TLBO) \citep{p24} works on the teaching and learning principle where knowledge flow from the better agents towards other. Each of the algorithms are applied on  mathematical representation of various kinds of applications and have been successful in achieving optimized solutions. 

\section{Nature of Green Heron Birds}
\label{sec-nature}
The Green Heron bird \citep{wiki} (Butorides virescens) resides on the freshwater or brackish water swampy marshes or wetlands with clumps of trees mainly in low lying areas where there are abundant scope of availability of fishes as their prey. They are nocturnal in habit and prefer to stay back in sheltered areas during the daytime. But when hungry they feed during the daytime. Their main food consists of small fish, spider, frogs, grasshoppers, snakes, rodents, reptiles, aquatic arthropods, mollusks, crustaceans, insects, amphibians, vertebrate or invertebrate animals like leeches and mice, provided they can catch.
\begin{figure*}[h!]
  \label{f1}
  \centering
    \includegraphics[width=0.5\textwidth]{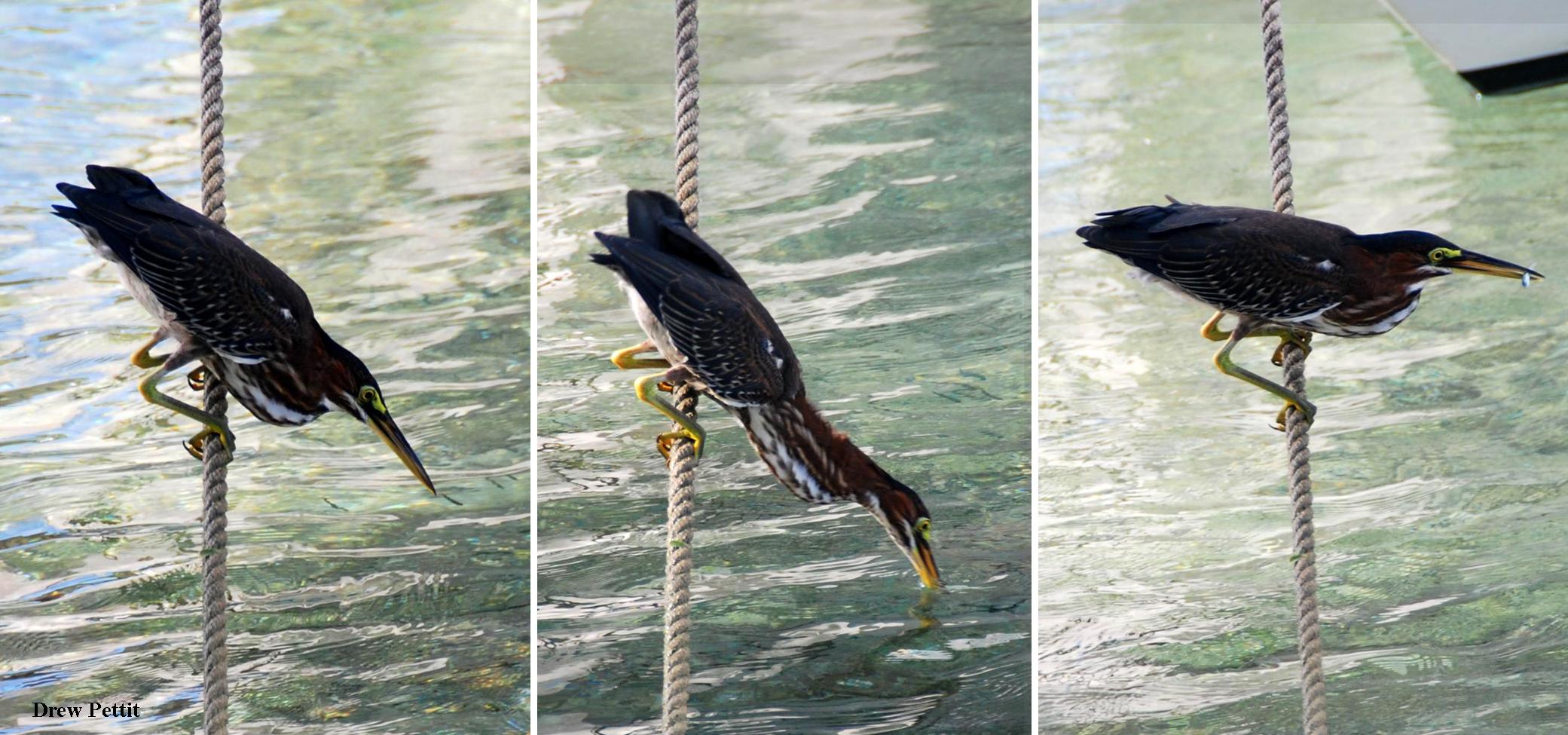}
  \caption{Preying Habit with Bait of Green Heron Bird}
\end{figure*}
Usually the Green Heron bird forages from a perch and there it stands with its body stretched out horizontally and lowered further, to insert its bill inside water for any unsuspecting prey. Green Heron is one among the few birds who can use tools for doing their daily jobs, the Green Heron will attract prey, mainly swarm of fishes, with bait (feathers, earthworms, bread crusts, tiny stick piece, insects, or even berries) when it drops on the water surface. The bait is dropped onto the water surface in order to attract fishes and all other water organisms that hover over the bait to sense its kind and food value. When any fish takes or tries to take the bait, the green heron bird grab hold the fish and eat the prey. This prey catching feature is being exploited as a meta-heuristic for complex problem solving and most importantly achieving optimization. The next section will describe in details the steps of the algorithm and its resemblance with the natural phenomenon of the Green Heron bird.

\section{Green Heron Optimization Algorithm Details}
\label{sec-GHSOA}
Overall the sequence of the Green Heron Optimization algorithm can be divided into the following basic operational steps which will perform different search based variations for a heuristic sequence or path generation and will thus establish a solution for the problems like graph based problems and combinatorial optimization problems, however this algorithm can be extended for discrete equations but with limited functionality of the algorithm. This is because in combinatorial optimization problem each of the individuals is a event whereas in equations each of the individuals are parameter values. However the individual constraints unique to each kind of problems are required to be established into the computation through implementation and the operations are just guidelines of what should be happening with the solution set individually or as a whole.

\begin{figure*}
\centering
\begin{minipage}{.5\textwidth}
  \centering
  \includegraphics[width=\textwidth]{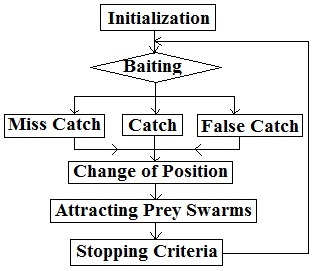}
  \captionof{figure}{Flow Diagram for Green Heron Swarm Optimization Algorithm}
  %\label{fig:test1}
  \label{f2}
\end{minipage}%
\begin{minipage}{.5\textwidth}
  \centering
  \includegraphics[width=\textwidth]{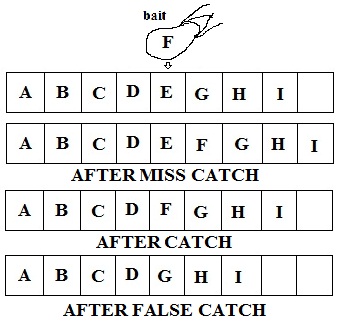}
  \captionof{figure}{Baiting Operation}
  \label{fobait}
  %\label{fig:test2}
\end{minipage}
\end{figure*}

%\begin{figure}[h!]
  %\label{fobait}
  %\centering
    %\includegraphics[width=0.5\textwidth]{ghoa}
  %\caption{Flow Diagram for Green Heron Swarm Optimization Algorithm}
%\end{figure}

\subsection{Baiting}
The baiting process is analogous to holding bait in the beak by the Green Heron bird, as it does and will drop at the appropriate place where there is chance of a catch or there are organisms in form of aquatic animals or fishes nearby. Similarly in the computation the bait is a solution subset that is arbitrarily generated (from a pool of such events and in case of constraints try avoiding the generation of already generated or most often generated events) and is held by the bird before it finds a good position through local search throughout the whole solution set. The bait and the prey is assumed as two individual solution subset which takes part in the operation and there are three alternatives which the bait-prey pair will bring about altering the solution set and thus heuristically creating a new solution set or improving it.  Now the three alternatives (the occurrences of which depends on the problem, its constraints, and the implementation and partly on the probability and local search) are:
\begin{itemize}
\item\textbf{MISS CATCH} - In this case the bait gets settled at one of its preferred place where it finds continuity and the bird fails to catch any prey and hence the number of solutions in the solution set tend to increase. Depending on the problem, the situation must be tackled. Like in Travelling Salesman Problems, scheduling problems, etc the missing node must be restored (may be from last, or at random) to sustain the validity of the solutions.
\item\textbf{CATCH} - In this case the bait helps the Green Heron bird to catch a prey and thus the solution set elements remains constant and one appropriate element is added and one inappropriate element is eliminated.
\item\textbf{FALSE CATCH} - Here the bird gets hold of a prey without using bait as sometimes fishes come near the surface of the water. In this step an inappropriate element of the solution set is eliminated from the set. The depletion of a node in the form of a catch form the solution set, must be restored for establishment of validity for certain constraints of the problems or limitation on the part of the variables.
\end{itemize}

%\begin{figure}[h!]
%\label{fobait}
 % \centering
  %  \includegraphics[width=0.5\textwidth]{obait}
  %\caption{Baiting Operation}
%\end{figure}

%\begin{figure}
%\centering
%\includegraphics[height=4cm]{obait}
%\caption{Baiting Operation.}
%\end{figure}
% Figure 3. Baiting Operation

In the Figure \ref{fobait}, the baiting operation of the Green Heron Optimization algorithm is demonstrated only for a particular case though many other cases can appear. Here we have a solution string consisting of A,B,C,D,E,G,H,I (where each one of them is an events or nodes or individual solution) and F is the bait which can operate on any position of the string. However the position is derived probabilistically and according to miss catch, catch and false catch (which can also be decided randomly or through Change of Position operational step, if possible) the three cases are shown.

For the problems like Travelling Salesman Problem and the discussed Quadratic Assignment Problem (QAP) and 0/1 Knapsack Problem (KSP), as the numbers of nodes are fixed and none should be repeated in the solution string, hence there must be replacement and of the replaced and deletion of the excess due to this phenomenon to compensate for the variation and to maintain the acceptability and validation of the solution set.

\subsection{Change of Position}
The Local search operation can occur through checking all (for small solution sets) or part (for long enough solution sets) of the solution sets for positions before it finds a suitable one. This step is analogous to the nature of the bird where it finds a suitable place where it can sit very near the surface of the water such that it can at any point of time can insert its beak inside water and take hold of a fish or aquatic animal whenever it comes near the surface naturally or due to the influence or temptation of the bait(s). This local search operation should count and made sure that too much time is not spend on a solution set if there are a number of solution sets to be taken care of in the iteration. In case of huge number elements, intensive local search strategy can be implemented for a selected zone to be checked or the low secondary fitness valued elements can be checked or any constraint of the problem can be utilized for such search and decision makings. In intensive local search the selected node is placed just by the side of the node where continuity can be established.

\begin{figure*}
\centering
\begin{minipage}{.5\textwidth}
  \centering
  \includegraphics[width=\textwidth]{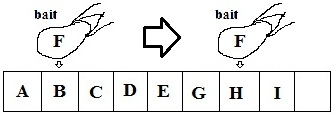}
  \captionof{figure}{Change of Position Operation}
  \label{f3}
  %\label{fig:test1}
\end{minipage}%
\begin{minipage}{.5\textwidth}
  \centering
  \includegraphics[width=\textwidth]{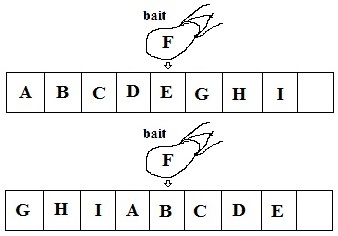}
  \captionof{figure}{Attracting Prey Swarms Operation}
  \label{fig:4}
  %\label{fig:test2}  
\end{minipage}
\end{figure*}

%\begin{figure}[h!]
  %\centering
    %\includegraphics[width=16pc, height=5pc]{oposition}
  %\caption{Change of Position Operation}
%\end{figure}

In the Figure 4 the Change of Position of Green Heron Optimization algorithm is shown. Here also we have considered the same solution string consisting of A,B,C,D,E,G,H,I and F is the bait which can operate on any position of the string. But the position is decided by the local search through a fractional portion of a particular location of the solution string. The best position is taken as the point of application. The best position depends on the local heuristic value, like in Travelling Salesman Problem it can be the least distance.

\subsection{Attracting Prey Swarms}
This Attracting Prey Swarms is also an equivalence of the local search operations that makes the algorithm quick convergent for constrained discrete problems solving the precedence criteria. But the step is little bit different from the Change of Position operation described previously. In this step the position of the bird sitting with the bait remains same but the swarm of fishes actually moves towards the fish or rather the bait is released and the fishes are attracted towards it. So for the solution set the point of release of the bait will remain same but the whole set will shift to create position for the best agent to receive the bait and this will help in an evolution like step where a shift can change the solution specially when the positions of the solution hold immense meaning and the correct sequence is of utmost importance. The example shown in Figure \ref{fig:4} will make the operation more clear. However this operation should be rare and occurs selectively for the iterations, only when there is no attachment in the initial positions of the solution set. This operation can be useful for problems like Travelling Salesman Problem, Vehicle Routing Problem, scheduling problems etc where the numbers of constraints are not present, but in problems like sequence ordering problems, routing, path planning, etc it can be useful selectively. However this operation can be operated on a selective portion of the solution set which is yet to be arranged or have not yet been lucky to engage in any kind of attachments. These local search processes are followed by the Baiting operation. The number in solution subset participating in shift is to be determined randomly and also the number of shifts must be less than the number in solution subset.
%\begin{figure}[h!]
  %\centering
    %\includegraphics[width=16pc, height=10pc]{oattracting}
  %\caption{Attracting Prey Swarms}
%\end{figure}

In the Figure 5, Attracting Prey Swarms operation of Green Heron Optimization Algorithm is demonstrated. In this step the position of bait is constant and the whole solution string or a selected portion of the string (depending on requirement and implementation) revolve to place an arbitrary node under it. Considering the same solution string consisting of A,B,C,D,E,G,H,I and F as the bait, E was under F, but thrice revolve will bring B under F. For Quadrature Assignment Problem and 0/1 Knapsack Problem we have considered the whole string to revolve.

%\begin{figure}
%\centering
%\includegraphics[height=4cm]{oattracting}
%\caption{Attracting Prey Swarms.}
%\end{figure}
% figure 5. Attracting Prey Swarms
It is to be mentioned that though the operations are described separately for convenience of understanding of the implementation with respect to any problem, in reality the operational steps are interweave and cannot be operated separately unless some problems may be suited for and the key lies in proper studding the problem into the algorithm.

\subsection{Brief Description of Fitness Function}
The fitness function used for determination and estimation of partial solution is quite important when it comes for the decision making of the system and optimization selection. It will help in a temporary deciding how well the nodes are placed and connected in a very well manner, however is valid for only one dimensional problems. But it is noticed that in majority graph based problems, obsessed with multi-objective optimization, the complete path is reached after a huge number of iterations and by the mean time it is very difficult to clearly demarcate the better incomplete result from the others and it is in this case the act of probabilistic steps can worsen the solution. Also the acts of the operations need to be operated in proper places mainly on the node gaps where there is yet to make any linkage. Hence a brief description of the secondary fitness function needs to be addressed. This secondary fitness value can be of several types: \\
1) The technique used in the simulation finds the linked consecutive nodes and is numbered with a number which denotes how many nodes are linked together at that portion. Say for example this can be regarded as an example.
\begin{equation}
  \text{Node Value} = \left\{		    \begin{array}{l l}
    			      					0 & ,\text{ for no linkage in either side} \\
    			      					1 & ,\text{ for linkage in one side} \\
    			      					2 & ,\text{ for linkage in both sides} 
  				    					\end{array} \right.	
\end{equation} 
Then the secondary fitness is calculated as (summation of the node value)/(number of nodes) that is with $N$ number of nodes we have, 
\begin{equation}
 \text{Secondary Fitness} = \frac{\sum\limits_{i=1}^{N} (Node Value)_i}{N}
\end{equation} 
High secondary fitness denotes that more numbers of nodes are linked together as a unit than the other solution string. But for the TSP as there occurs a link between every node, the secondary fitness will always be constant and will be of no use. The solution with maximum secondary fitness will be the better solution. In the simulation this procedure is used. \\
2) Another partial solution fitness evaluation can be through the use of the number of partial solution that is linked portion are present in that string which can have high probability of being processed into a complete path than the isolated ones. Here on the count of the linked sections present in the solution string are kept as secondary fitness value. But contrary to the previous method, this method provides minimum as best result.
Say in a solution string represented as $S = \{a_1,a_2,\ldots,a_N\}$ and the lined portions are represented as $\{b_1,b_2,\ldots,b_m\}$ where each $b_i \subseteq S$ and $i \in \{1,2,\ldots,m\}$ and 
\begin{equation}
  \text{Secondary Fitness} = m
\end{equation}
and with formation of linkages $m$ will constantly decrease. In this case we can use each of the subsets as nodes for link foundation. 

\subsection{Adaptiveness of GHSOA}
The Green Heron Swarm Optimization Algorithm provides adaptive scheme for variable handling for each data set and can be used where the dimension is not constant like in path planning, adaptive clustering, unsupervised learning based clustering etc. It is good for experimentation for the problem which can be multi-variable during the initial positions and gradually varies with iterations. Also for constant length combinatorial optimization, it can be modified accordingly and can be regarded as a special case of the algorithm. There can be situations when the linkage between two path segments can be done through single or multiple numbers of nodes and this requires the need of adaptive flexibility in the operators. For adaptive clustering, this algorithm can be helpful for generation and maintenance of the cluster centroids.

\subsection{Location Based Neighbour Influenced Variation (LBNIV)}
Location Based Neighbour Influenced Variation (LBNIV) is the adaptive variation scheme for the continuous domain problems like the numerical benchmark equations and mathematical models of applications. It follows the habit of the bird to get influence by the different elements of the environment mainly water organisms and then act accordingly for better catching availability and better attraction of swarms of fishes. Hence this equation is called as being influenced by the neighbours who have better position or from where there can be availability of better opportunity. For each continuous variable of the fitness function of the problem that is $x_t \in \{x1_t,x2_t,\ldots,xD_t \}$ where $D$ is the dimension of the problem and $xD_t$ is the $D^{th}$ element in the $t^{th}$ iteration. We have the following equations governing the variation of the variable's variation and is self-adaptive, previous iteration based and error sensitive based on the fitness function of the $t$ and $t-1$ iteration.

\begin{equation}
  \label{e4}
  x_t = x_{t-1} + |(x_{best} - x_r)|d_{(t,r)} \epsilon_{(t,r)} + |(x_{best} - x_f)|d_{(t,f)} \epsilon_{(t,f)} + bias
\end{equation}
where we have $[\{d_{(t,r)}$, $d_{(t,f)}\} \in d_{t}]$ and $[\{\epsilon_{(t,r)},\epsilon_{(t,f)}\} \in \epsilon_{t}]$ given by $d_{(t+1)}$ and $\epsilon_{(t+1)}$ of Equation \ref{e1} and Equation \ref{e2} respectively of previous iteration as
\begin{equation}
  \label{e1}
  d_{(t+1)}   = \left\{ \begin{array}{ll}
		          \frac{J_{t-1} - J_{t}}{|J_{t-1}|} &\mbox{ for $x_t \geq x_{t-1}$} \\
                  \frac{J_{t} - J_{t-1}}{|J_{t-1}|} &\mbox{ for $x_t < x_{t-1}$}
                 \end{array} \right.
\end{equation}
\begin{equation}
  \label{e2}
  \epsilon_{(t+1)} = \left\{ \begin{array}{ll}
		          \epsilon_{t}/k &\mbox{ for $x_t > x_{max}$} \\
                  \epsilon_{t}*k &\mbox{ for $x_t < x_{min}$} \\
                  \epsilon_{t} &\mbox{ Otherwise}
                 \end{array} \right.
\end{equation}
for minimization problem with $J_t < J_{t-1}$ or $J_t > J_{t-1}$ where $xi_t$ is the $i^{th}$ variable for iteration $t$, $J_t$ is the fitness value at iteration $t$, $x_f$ and $x_r$ are the front and rear neighbours respectively where $xD_f \in \{xD_t \text{ of Next Fellow Agents } \}$ ad $x_r \in \{xD_t \text{ of Previous Fellow Agents } \}$, $\delta_t$ is the variations contributor with respect to the $t^{th}$ iteration generated from previous iteration $t-1$, $\epsilon$ is the constantly changing adaptive contributor which denotes what percentage of the $\delta$ must be incorporated into the variable and depends on the bound restriction of the variable. For $\epsilon = 1$, the value of $\delta$ is same for all the increasing or decreasing variables, but if the boundary of $xi_t$ varies then we have separate $\epsilon$ for different $xi_t$ variables. It is better to keep $\delta$ set and $\epsilon$ set separately for each variable and thus in total for $D$ variables for each solution set. It is to be noticed here that the Equation \ref{e1} and Equation \ref{e2} always tries to drag the value of $xi_t$ towards the best that is for minimum of $J_t$ for minimal optimization. 

\section{Performance Evaluation}
\label{sec-performance}
Green Heron Swarm Optimization Algorithm was itself developed for discrete problems but is tested on continuous numerical benchmarks. The details of the algorithmic steps for GHSOA for the various problems are provided below.

\subsection{GHOSA for Travelling Salesman Problem (TSP) / Quadrature Assignment Problem (QAP)}
Table 1 and Table 2 have provided the results of datasets set \citep{s1},\citep{s2} for Travelling Salesman Problem and Quadrature Assignment Problem respectively and compared with optimum value. GHOSA was operated on each of them.

The Algorithm for GHSOA for Travelling Salesman Problem / Quadrature Assignment Problem is being provided.\\
\textbf{Step 1:} Initialize the Solution set all its n (n = dimension of problem) events $\{x_1,x_2,\ldots,x_n \}$.\\  
\textbf{Step 2:} Generate N Solution set each consisting of $\{x_1,x_2,\ldots,x_n \}$ where each element denote an event and the corresponding position in the array denotes a fixed positional significance. \\
\textbf{Step 3:} Evaluate the fitness of each string, Store the value of profit if constraint satisfied else make it zero. Update the Global best if better is found.\\
\textbf{Step 4:} (For each string) Perform ``Baiting'' (with Miss Catch, Catch, False Catch) where position is selected randomly (need to take care of the duplicates) or at selected points depending upon implementation on deterministic approach or probability. \\
\textbf{Step 5:} Perform ``Change of Position'' operation depending upon the requirement and initial search results. (Random positioning is done to see which combination yield best result) Here on selected or the whole string depending on the pseudo-random generation of the two operation parameters.\\
\textbf{Step 6:} Perform ``Attracting Prey Swarms''.  Complete ``Baiting'' operation. \\
(End of For each string) \\
\textbf{Step 7:} Evaluate the fitness of each string. Store the value of profit if constraint satisfied else make it zero. Update the Global best if required. If New (derived out of combination of operation(s)) is better, then replace the old else don't. The global best consists of the fitness value along with the string consisting of 1s and 0s. \\
\textbf{Step 8:} After each iteration replace X\% worst solutions with random initialization. (X depends on N and according to the exploration requirement) \\
\textbf{Step 9:} If number of iteration is complete then stop else continue from Step 4.

\scriptsize
\begin{longtable}{|c|c|r||c|c|c|c|c|}
\caption{Evaluation of EVSOA on TSP Datasets}\\
\hline
\textbf{Name} & \textbf{Dim} & \textbf{Optimum} & \textbf{Mean} & \textbf{SD} & \textbf{Best} & \textbf{Worst}\\
\hline
\endfirsthead
\multicolumn{8}{c}%
{\tablename\ \thetable\ -- \textit{Continued from previous page}} \\
\hline
\textbf{Name} & \textbf{Dim} & \textbf{Optimum} & \textbf{Mean} & \textbf{SD} & \textbf{Best} & \textbf{Worst}\\
\hline
\endhead
\hline \multicolumn{8}{r}{\textit{Continued on next page}} \\
\endfoot
\hline
\endlastfoot
 Ulysses16.tsp &16  & 	74.11 		& 	75.18 		& 0.047 & 	74.11 		& 78.27 		\\\hline
 att48.tsp	 & 	48  & 	3.3524e+004 & 	3.5436e+004	& 12.1  & 	3.3613e+004	& 4.2996e+004 	\\\hline
 st70.tsp	 & 	70  & 	678.5975 	& 	711.676 	& 115.9 & 	694 		& 746   		\\\hline
 pr76.tsp	 & 	76  & 	1.0816e+005 & 	1.3319e+005 & 125.7 & 	1.0816e+005 & 1.3757e+005 	\\\hline
 gr96.tsp	 & 	96  & 	512.3094 	& 	643.97 		& 69.4  & 	573.16 		& 806.4 		\\\hline
 gr120.tsp	 & 	120 & 	1.6665e+003 & 	1.7963e+003 & 46.8  & 	1.7112e+003 & 1.8753e+003 	\\\hline
 gr202.tsp	 & 	202 & 	549.9981 	& 	839.19	 	& 178.2 & 	610.8 		& 1005.9 		\\\hline
 tsp225.tsp	 & 	225 & 	3919	 	& 	4151.8 		& 213.7 & 	4058.9 		& 5034.8 		\\\hline
 a280.tsp	 & 	280 & 	2.5868e+003 & 	2.77913e+003 & 986.1 & 	2.6772e+003 & 3.1463e+003 	\\\hline
\end{longtable}

\begin{longtable}{|c|c|r||c|c|c|c|c|}
\caption{Evaluation of EVSOA on QAP Datasets}\\
\hline
\textbf{Name} & \textbf{Dim} & \textbf{Optimum} & \textbf{Mean} & \textbf{SD} & \textbf{Best} & \textbf{Worst} & \textbf{Error}\\
\hline
\endfirsthead
\multicolumn{8}{c}%
{\tablename\ \thetable\ -- \textit{Continued from previous page}} \\
\hline
\textbf{Name} & \textbf{Dim} & \textbf{Optimum} & \textbf{Mean} & \textbf{SD} & \textbf{Best} & \textbf{Worst} & \textbf{Error}\\
\hline
\endhead
\hline \multicolumn{8}{r}{\textit{Continued on next page}} \\
\endfoot
\hline
\endlastfoot
  chr15a & 15 & 9552     & 9552     & 0 	 & 9552 & 9552 & 0				\\\hline
  bur26a & 26 & 5426670  & 5514119  & 8990.6 & 5426670 & 5512133 & 1.6115 	\\\hline
  chr12a & 12 & 9552     & 9552     & 0 	 & 9552 & 9552 & 0 				\\\hline
  chr18a & 18 & 11098    & 11098    & 0 	 & 11098 & 11098 & 0			\\\hline
  chr20a & 20 & 2192     & 2372.5   & 72.1   & 2192 & 2561 & 8.2345			\\\hline
  chr22a & 22 & 6156     & 6525.4   & 90.92  & 6156 & 6502 & 6.0006			\\\hline
  chr25a & 25 & 3796     & 4231.9   & 300.2  & 3796 & 7927 & 11.4831		\\\hline
  els19  & 19 & 17212548 & 17212548 & 0 	 & 17212548 & 17212548 & 0 		\\\hline
  esc16a & 16 & 68 		 & 68       & 0 	 & 68 & 68 & 0					\\\hline
  esc32e & 32 & 2 		 & 2 		& 0 	 & 2 & 2 & 0					\\\hline
  had14  & 14 & 2724     & 2724 	& 0 	 & 2724 & 2724 & 0				\\\hline
  nug24  & 24 & 3488     & 3795 	& 92.98  & 3488 & 3729 & 8.8016			\\\hline
  nug27  & 27 & 5234     & 5416.4 	& 121.75 & 5234 & 5518 & 3.4849			\\\hline
  esc16b & 16 & 292      & 292 		& 0 	 & 292 & 292 & 0				\\\hline
  nug16a & 16 & 1610     & 1610 	& 0 	 & 1610 & 1610 & 0				\\\hline
  nug20  & 20 & 2570     & 2570 	& 0 	 & 2570 & 2570 & 0				\\\hline
 \hline
\end{longtable}
\normalsize

\subsection{GHOSA for 0/1 Knapsack Problem (KSP)}
In this part the dataset of 0/1 Knapsack Problem \citep{p3} is being operated with GHSOA for optimization with respect to optimum value.

The Algorithm for GHSOA for 0/1 Knapsack Problem .\\
\textbf{Step 1:} Consider a dataset of $n$ items and $x_i$ is the parameter to denote the inclusion or exclusion of the $i^{th}$ item for any bag, for $m$ number of bags where each bag is has maximum capacity $W_m$.\\
\textbf{Step 2:} Generate $N$ solution strings for each type of bag \textbf{m} where each string has $x_i$ for $i = \{1,2,\ldots,n\}$ consisting of n items and each is represent by a number from 1 to n without repetition of any of the numbers in the string.\\
\textbf{Step 3:} Now generate a threshold between 1 and n in integer values, so that the string of integer values can be converted to string of 0s and 1s and the random value of threshold will decide how many of them should be 0s and 1s and also the positional values will create combinations and due to the GHOA the integer values for each positions will change. \\
\textbf{Step 4:} For $(x_i >$ threshold) Make it 1, else Make it 0. So string of 0 \& 1 represent $x_i$ set.\\
\textbf{Step 5:} Evaluate the fitness of each string, Store the value of profit if constraint satisfied else make it zero. Update the Global best if required. Convert the \textbf{x} vector from 0/1 values back to its previous form of numbers from 1 to n.\\
(For each string) \\
\textbf{Step 6:} Perform ``Baiting'' (with Miss Catch, Catch, False Catch) where position is selected randomly (need to take care of the duplicates) or at selected points depending upon implementation on deterministic approach or probability. \\
\textbf{Step 7:} Perform ``Change of Position'' operation depending upon the requirement and initial search results. (Random positioning is done to see which combination yield best result) Here on selected or the whole string depending on the pseudo-random generation of the two operation parameters.\\
\textbf{Step 8:} Perform ``Attracting Prey Swarms''.  Complete ``Baiting'' operation. \\
(End of For each string) \\
\textbf{Step 9:} Repeat Step 3 and 4 to make the solution strings eligible for fitness evaluation. Now Evaluate the fitness of each string. Store the value of profit if constraint satisfied else make it zero. Update the Global best if required. If New (derived out of combination of operation(s)) is better, then replace the old else don't. The global best consists of the fitness value along with the string consisting of 1s and 0s. \\
\textbf{Step 10:} After each iteration replace $X$\% worst solutions with random initialization. ($X$ depends on $N$ and according to the exploration requirement) \\
\textbf{Step 11:} If number of iteration is complete then stop else continue from Step 3.

\scriptsize
\begin{longtable}{|c|c|r||c|c|c|c|c|}
\caption{Evaluation of EVSOA on 0/1 Knapsack Datasets}\\
\hline
\textbf{Name} & \textbf{Dim} & \textbf{Optimum} & \textbf{Mean} & \textbf{SD} & \textbf{Best} & \textbf{Worst} & \textbf{Error}\\
\hline
\endfirsthead
\multicolumn{8}{c}%
{\tablename\ \thetable\ -- \textit{Continued from previous page}} \\
\hline
\textbf{Name} & \textbf{Dim} & \textbf{Optimum} & \textbf{Mean} & \textbf{SD} & \textbf{Best} & \textbf{Worst} & \textbf{Error}\\
\hline
\endhead
\hline \multicolumn{8}{r}{\textit{Continued on next page}} \\
\endfoot
\hline
\endlastfoot
  WEISH01 & 5,30  & 4554   & 4463.2   & 23.4789  & 4549   & 4423   & 1.9939	\\\hline
  WEISH07 & 5,40  & 5567   & 5325.1   & 92.723   & 5493   & 5013   & 4.3452	\\\hline
  WEISH10 & 5,50  & 6339   & 5865.842 & 184.8216 & 6309   & 5592   & 7.4642	\\\hline
  WEISH15 & 5,60  & 7486   & 6918.92  & 213.202  & 7332   & 6819   & 7.5752	\\\hline
  WEING8  & 2,105 & 624319 & 552021   & 4000.2   & 610322 & 529391 & 11.5803\\\hline
  WEING1  & 2,28  & 141278 & 141156.1 & 578.4875 & 141278 & 141010 & 0.0863	\\\hline
  FLEI    & 10,20 & 2139   & 2139     & 0        & 2139   & 2139   & 0		\\\hline
  HP1     & 4,28  & 3418   & 3406     & 45.02    & 3418   & 3293   & 0.3511	\\\hline
  PB6 	  & 30,40 & 776    & 728.1    & 11.2038  & 756    & 701    & 6.1727	\\\hline
  PET2 	  & 10,10 & 87061  & 87061    & 0        & 87061  & 87061  & 0		\\\hline
  PET3    & 10,15 & 4015   & 4015     & 0        & 4015   & 4015   & 0		\\\hline
  PET4    & 10,20 & 6120   & 6120     & 0        & 6120   & 6120   & 0		\\\hline
  PET5    & 10,28 & 12400  & 12326.74 & 52.954   & 12400  & 12129  & 0.5908	\\\hline
  PET6    & 5,39  & 10618  & 10499.87 & 32.982   & 10570  & 10446  & 1.1125	\\\hline
  PET7    & 5,50  & 16537  & 16336    & 22.956   & 16393  & 16078  & 1.2155	\\\hline
\end{longtable}
\normalsize

%\subsection{GHOSA for Sequential Ordering Problem}
%\subsection{GHOSA for Resource Constrained Shortest Path Problem}

\subsection{GHOSA for Multi-Objective Road Network / Resource Constrained Shortest Path Problem (RCSP)}
\begin{figure*}
\centering
\begin{minipage}{.5\textwidth}
  \centering
  \includegraphics[width=\textwidth]{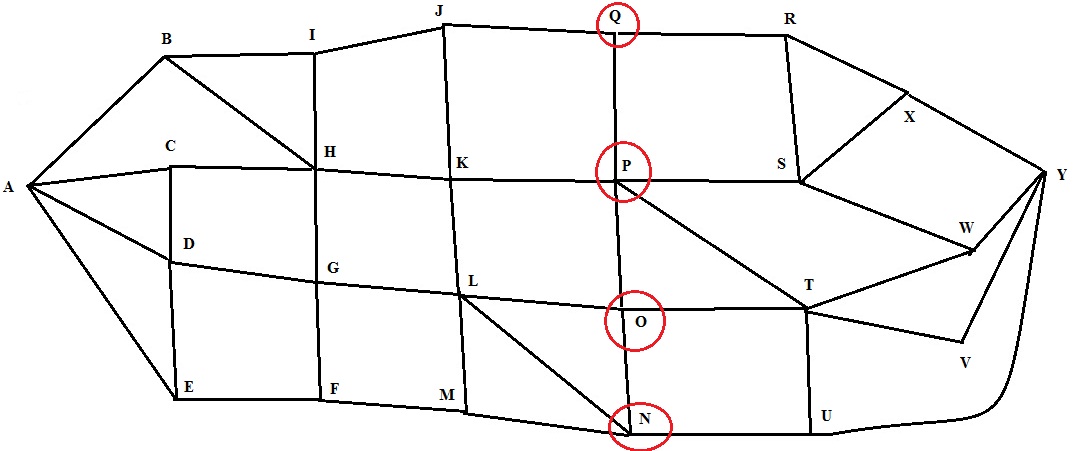}
  \captionof{figure}{Road Graph Network}
  \label{f5}
\end{minipage}%
\begin{minipage}{.5\textwidth}
  \centering
  \includegraphics[width=\textwidth]{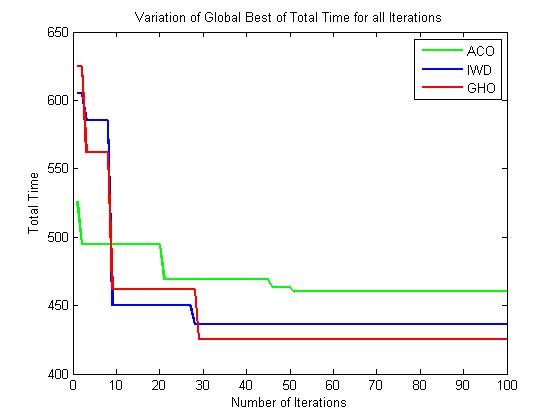}
  \captionof{figure}{Variation of Global Best of Total Time for all Iterations}
  \label{f6}
\end{minipage}
\end{figure*}

The road network considered here is shown in Figure \ref{f5} which is being measured to see approximately how the optimized vehicle route flow in the network. Each of the edges is provided with a distance and an average waiting time assumed to be calculated on the basis of the data collected independently by the sensor network present at that crossing. The equations for distance and waiting time for the road network which is required to be minimized are two summation based equation which depends on the path traversed by the agents on its way from the source A to destination Y.
\begin{equation}
\label{e5}
 f_1 = \sum\limits_{k=1}^{n} \frac{D_k}{V}
\end{equation}
\begin{equation}
\label{e6}
 f_2 = \sum\limits_{k=1}^{n} (AWT)_k
\end{equation}
where $f_1$ and $f_2$ are the two equations and $D_k$ is the distance and $(AWT)_k$ is the average waiting time for the path $\{k \subseteq \{i,j\} \in G\}$, $G$ is the graph network. The fitness function $f$ considered is $f = (f_1+f_2)$. $V$ is the constant velocity for normalization of the distance and it makes $f_1$ same as $f_2$ in unit of time.

There are a few assumptions made on the behalf of the road model so as to make it simple and avoid unnecessary details and at the same time make it acceptable for the simulation and the algorithm. The vehicles considered here are uniform in size, speed and non-accelerating. The time taken for movement and waiting are crisp and other details like size of queue before the vehicle, width of the road etc are all discarded and an average of the all the happenings are considered. The model emphasis on the movement of the vehicles from the source to destination and in the meantime only these vehicles are considered or rather accounted for the simulation and conclusion generation. However other vehicles which are also present and are destined to move from other parts of the network graph to some other places are not considered in details but their presence is established through some random variation of the parameters like waiting time, queue length etc.

The Algorithm for GHSOA for road network and similar other problems like Resource Constrained Shortest Path Problem (RCSP) is given below.\\
Step 1: Initialize the equation and all its n (n = dimension) variables as $\{x_1,x_2,\ldots ,x_n\}$ \\
Step 2: Initialize $N$ strings each consisting of $\{x_1,x_2,\ldots ,x_n\}$ as coefficients having random numerical values. (evaluate the constraints and bounds of the variables and reinitialize the strings which are violating them). [positions are related to specific variables and are fixed] \\
Step 3: Initialize the fitness matrix and Evaluate the fitness of each string and set global best result. \\
Step 4: (For each string) Perform ``Baiting'' (with Miss Catch, Catch, False Catch) where position is selected with some intensive local search strategy where the nodes with least secondary fitness is searched (probability, random partial string, etc strategies can also be used). \\
Step 5: Perform ``Change of Position'' depending upon the requirement and initial search results. \\
Step 6: Perform ``Attracting Prey Swarms''.   \\
Step 7: Complete ``Baiting'' operation. \\
Step 8: Perform ``Location Based Neighbour Influenced Variation'' for each of $\{x_1,x_2,\ldots ,x_n\}$ (with $\epsilon = 1$ initially but gradually changes). (End of For each string) \\
Step 9: Evaluate the fitness of each string considering the validity (boundation and constraints) of each $x_i$ where $x_i \in \{x_1,x_2,\ldots ,x_n\}$. \\
Step 10: If New is better, then replace the old else don't. \\
Step 11: Select the best result and compare with global best. If better then set it as global best. \\
Step 12: After each iteration replace X\% worst solutions with random initialization. \\
Step 13: If number of iteration is complete then stop else continue from Step 4.

% insert image for "GHOSA for Road Network, Compared with ACO"

\begin{figure*}
\centering
\begin{minipage}{.5\textwidth}
  \centering
  \includegraphics[width=\textwidth]{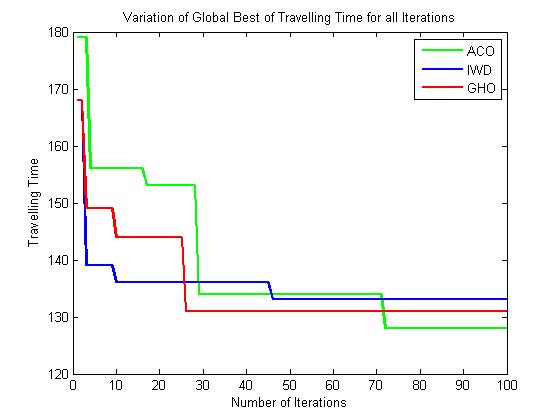}
  \captionof{figure}{Variation of Global Best of Travelling Time for all Iterations}
  \label{f7}
\end{minipage}%
\begin{minipage}{.5\textwidth}
  \centering
  \includegraphics[width=\textwidth]{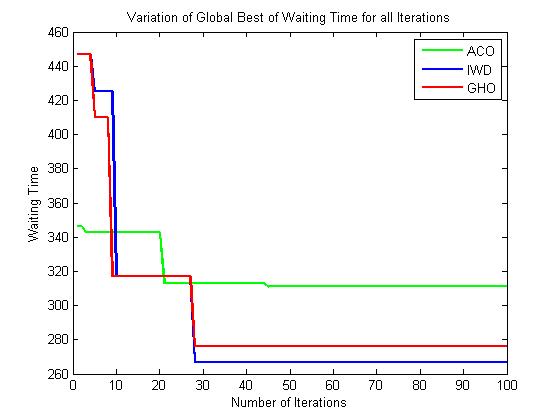}
  \captionof{figure}{Variation of Global Best of Waiting Time for all Iterations}
  \label{f8}
\end{minipage}
\end{figure*}

\begin{figure*}
\centering
\begin{minipage}{.5\textwidth}
  \centering
  \includegraphics[width=\textwidth]{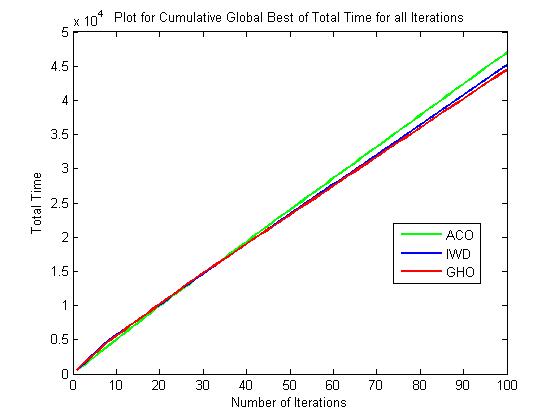}
  \captionof{figure}{Plot for Cumulative Global Best of Total Time for all Iterations}
  \label{f9}
\end{minipage}%
\begin{minipage}{.5\textwidth}
  \centering
  \includegraphics[width=\textwidth]{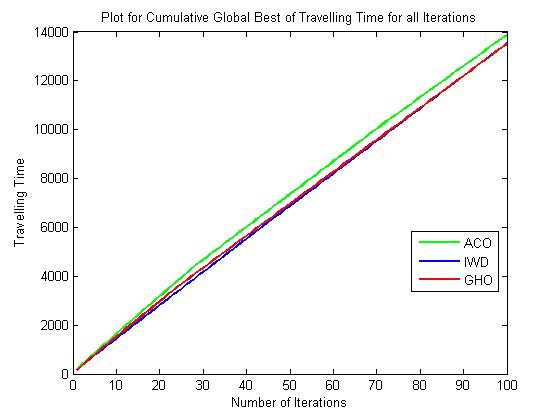}
  \captionof{figure}{Plot for Cumulative Global Best of Travelling Time for all Iterations}
  \label{f10}
\end{minipage}
\end{figure*}

\begin{figure*}
\centering
\begin{minipage}{.5\textwidth}
  \centering
  \includegraphics[width=\textwidth]{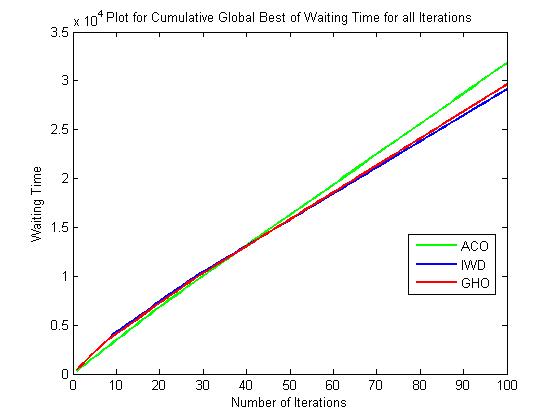}
  \captionof{figure}{Plot for Cumulative Global Best of Waiting Time for all Iterations}
  \label{f11}
\end{minipage}%
\begin{minipage}{.5\textwidth}
  \centering
  \includegraphics[width=\textwidth]{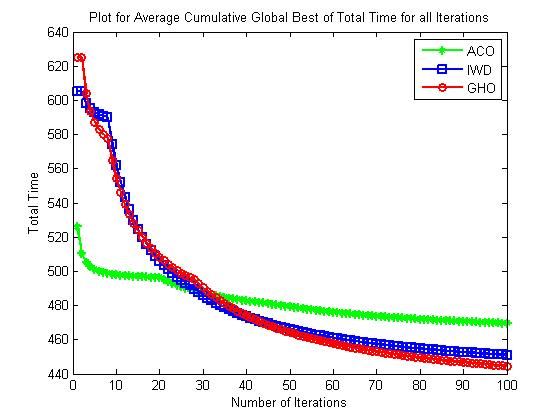}
  \captionof{figure}{Plot for Average Cumulative Global Best of Total Time for all Iterations}
  \label{f12}
\end{minipage}
\end{figure*}

\begin{figure*}
\centering
\begin{minipage}{.5\textwidth}
  \centering
  \includegraphics[width=\textwidth]{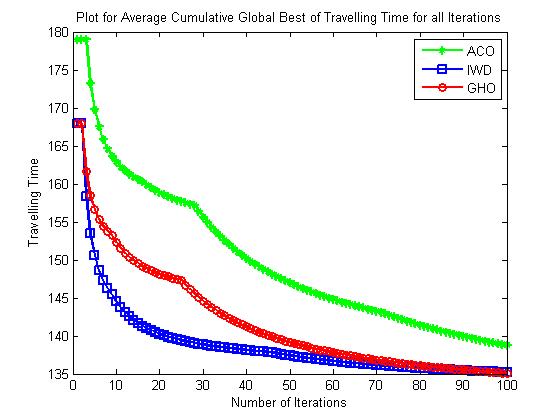}
  \captionof{figure}{Plot for Average Cumulative Global Best of Travelling Time for all Iterations}
  \label{f13}
\end{minipage}%
\begin{minipage}{.5\textwidth}
  \centering
  \includegraphics[width=\textwidth]{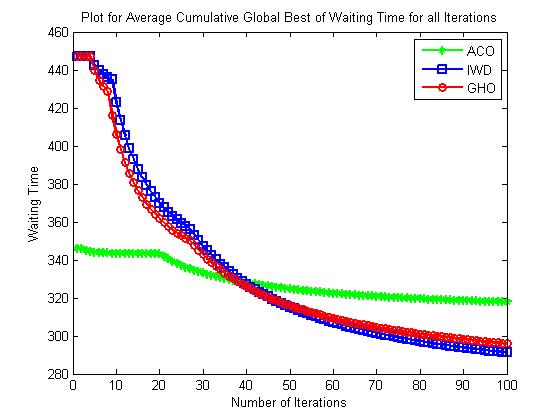}
  \captionof{figure}{Plot for Average Cumulative Global Best of Waiting Time for all Iterations}
  \label{f14}
\end{minipage}
\end{figure*}

\subsection{GHOSA for Benchmark Equations}
Numerical Benchmark Equations (along with dimension, optimized value) given in Table 4 is used for simulation of GHSOA and the corresponding results (in the form of mean, standard deviation, best, worst, error) are provided in Table 5. In this section we have described the details of implementation of the algorithm along with the addition of the location based neighbour influenced variation scheme in Step 8 and problem oriented variations of the GHSOA for the benchmark equations: \\
Step 1: Initialize the equation and all its n (n = dimension) variables as $\{x_1,x_2,\ldots ,x_n\}$ \\
Step 2: Initialize $N$ strings each consisting of $\{x_1,x_2,\ldots ,x_n\}$ as coefficients having random numerical values. (evaluate the constraints and bounds of the variables and reinitialize the strings which are violating them). [positions are related to specific variables and are fixed] \\
Step 3: Initialize the fitness matrix and Evaluate the fitness of each string and set global best result. \\
Step 4: (For each string) Perform ``Baiting'' (with Miss Catch, Catch, False Catch) where position is selected with some intensive local search strategy where the nodes with least secondary fitness is searched (probability, random partial string, etc strategies can also be used). \\
Step 5: Perform ``Change of Position'' depending upon the requirement and initial search results. \\
Step 6: Perform ``Attracting Prey Swarms''.   \\
Step 7: Complete ``Baiting'' operation. \\
Step 8: Perform ``Location Based Neighbour Influenced Variation'' for each of $\{x_1,x_2,\ldots ,x_n\}$ (with $\epsilon = 1$ initially but gradually changes). (End of For each string) \\
Step 9: Evaluate the fitness of each string considering the validity (boundation and constraints) of each $x_i$ where $x_i \in \{x_1,x_2,\ldots ,x_n\}$. \\
Step 10: If New is better, then replace the old else don't. \\
Step 11: Select the best result and compare with global best. If better then set it as global best. \\
Step 12: After each iteration replace X\% worst solutions with random initialization. \\
Step 13: If number of iteration is complete then stop else continue from Step 4.

% insert the benchmark table here
\subsection{Details of Benchmarks}
Several famous benchmark equations and its description, constraints (in form of range values for each variable) and optimized values are being provided in Table 2 which are used for the simulation of the Egyptian Vulture Swarm Optimization Algorithm to see how it performs under various dimensions and range constraints of the benchmark equations. These equations are standardized famous mathematical representations that require optimization and are used as test-beds for analysis of the algorithms. The list of the benchmark equations are provided in Table X. They are Sphere Function $(f_1)$, Rosenbrock Function $(f_3)$, Hump Functions $(f_6)$, Branin Function $(f_7)$, Goldstein \& Price Function $(f_8)$, Power Sum Function $(f_9)$, Beale Function $(f_{10})$, Colville Function $(f_{11})$, Sum Squares Function $(f_{12})$, Dekkar and Aarts $(f_{13})$, McCormick $(f_{14})$, Two Peak Trap  $(f_{15})$, Central Two Peak Trap $(f_{16})$, Five uneven peak Trap $(f_{17})$, Equal Maxima $(f_{18}$, Decreasing Maxima $(f_{19})$, Uneven Maxima $(f_{20})$, Uneven Decreasing Maxima  $(f_{21})$, Himmelblau's function $(f_{22})$, Six-hump Camel Back function $(f_{23})$, Michalewics Function $(f_{24})$, Matyas Function $(f_{25})$.

The computational result of simulation of the Green Heron Swarm Optimization Algorithm on the benchmark problems is performed on Matlab R2011a (in a system with configuration of Core i3-2330M 2.20 GHz Intel 2nd Generation Processor and 4 GB RAM) and is provided in the Table \ref{Tsol} having the equation ID (corresponding to Table \ref{Teq}), dimension of the equations or considered dimension, mean, standard deviations (SD), best solution, worst solution and mean error. The computational results, compared with the traditional Genetic Algorithm (GA) and Particle Swarm Optimization (PSO), clearly revealed how the algorithm worked for the maximum 25000 iterations mark and has revealed great convergence for the solutions. Solution and convergence does not depend on the initial value of $\epsilon_t$ which is kept as usual quite low with 0.2 and k=2 and bias=0.001.

% table for travelling salesman problem
% table for knapsack
% table for quadratic assignemnt problem
% table for benchmark equations
% table for benchmark results

\scriptsize
\begin{longtable}{|p{0.3cm}|p{5cm}|p{0.3cm}|p{2cm}|p{1.3cm}|}
\caption{Details of Benchmark Equations} \label{Teq}\\
\hline
\textbf{\#} & \textbf{Equation} & \textbf{D} & \textbf{Range} & \textbf{Optimum}\\
\hline
\endfirsthead
\multicolumn{5}{c}%
{\tablename\ \thetable\ -- \textit{Continued from previous page}} \\
\hline
\textbf{\#} & \textbf{Equation} & \textbf{D} & \textbf{Range} & \textbf{Optimum}\\
\hline
\endhead
\hline \multicolumn{5}{r}{\textit{Continued on next page}} \\
\endfoot
\hline
\endlastfoot
 $f_1$ &    $\sum\limits_{i=1}^{D}x_i^2$  & $10$  & $[-20,20]_D$ & 0 \\
 \hline
 $f_2$ &    $\sum\limits_{i=1}^{D}|x_i|+\prod\limits_{i=1}^{D}|x_i| $  & $10$  & $[-20,20]_D$ & 0  \\
 \hline
 $f_3$ &    $\sum\limits_{i=1}^{D-1}[100(x_{i+1}-x_i^2)^2 + (x_i-1)^2]$  & $10$  & $[-20,20]_D$ & 0  \\
 \hline
 $f_4$ &    $\sum\limits_{i=1}^{D}(x_i - 0.5)^2$  & $10$  & $[-20,20]_D$ & 0  \\
 \hline
 $f_5$ &    $\sum\limits_{i=1}^{D}[x_i^2 -10cos(2\pi x_i) + 10]$  & $10$  & $[-5.12,5.12]_D$ & 0  \\
 \hline
 $f_6$ &    $4x_1^2 - 2.1x_1^4 + \frac{1}{3}x_1^6 + x_1x_2 - 4x^2_2 + 4x^4_2 $  & $2$  & $[-5,5]_D$ & -1.03163  \\
 \hline
 $f_7$ &    $(x_2 - \frac{5.1}{4\pi^2}x_1^2 + \frac{5}{\pi}x_1- 6)^2 + 10(1 - \frac{1}{8\pi})cosx_1 + 10$  & $2$  & $[-5,10]x[0,15]$ & .398  \\
 \hline
 $f_8$ &    $   [1+(x_1+x_2+1)^2(19-14x_1+3x_1^2
 							   -14x_2+6x_1x_2+3x_2^2)]*[30+(2x_1
 							   -3x_2)^2*(18-32x_1+12x_1^2+48x_2-36x_1x_2+27x_2^2)]   $  & $2$  & $[-2,2]_D$ & 3  \\
 \hline
 $f_9$ &    $\sum\limits_{i=1}^D (\sum\limits_{j=1}^i x_j)^2$  & $10$  & $[-20,20]_D$ & 0  \\
 \hline
 $f_{10}$ & $[1.5-x_1(1-x_2)]^2+[2.25-x_1(1-x_2^2)]^2+[2.625-x_1(1-x_2^3)]^2 $  & $2$  & $[-4.5,4.5]_D$ & 0  \\
 \hline
 $f_{11}$  &  $ 100[x_2-x_1^2]^2 + (1-x_1)^2 + 90(x_4 - x_3^2)^2 + (1-x_3)^2 + 10.1[(x_2-1)^2+(x_4-1)^2] + 19.8(x_2-1)(x_4-1)  $ & $4$  & $[-10,10]_D$ & 0  \\
\hline
 $f_{12}$   & $\sum\limits_{i=1}^{D}ix_i^4 + random[0,1)$  & $10$  & $[-1.28,1.28]_D$ & 0 \\
 \hline
 $f_{13}$   & $10^5x_1^2 + x_2^2 - (x_1^2 + x_2^2)^2 + 10^{-5}(x_1^2 + x_2^2)^4$  & $2$  & $[-20,20]_D$ & -24777  \\
 \hline
 $f_{14}$   & $sin(x_1 + x_2) + (x_1 - x_2)^2 -\frac{3}{2}x_1+\frac{5}{2}x_2 + 1$ & $2$  & $ \begin{matrix}[-1.5,4]x \\ [-3,3]\end{matrix}$ & -1.9133  \\
 \hline
 $f_{15}$   & $ \left\{ \begin{array}{ll}
		          \frac{160}{15}(15-x) &\mbox{ for $0\leq x <15$} \\
                  \frac{200}{5}(x-15) &\mbox{ for $15\leq x \leq 20$}
                 \end{array} \right. $    & $1$  & $[0,20]$ & 0  \\
 \hline
 $f_{16}$  & $  \left\{ \begin{array}{ll}
		          \frac{160}{10}x &\mbox{ for $0\leq x <10$} \\
                  \frac{160}{5}(15-x) &\mbox{ for $10\leq x <15$} \\
                  \frac{200}{5}(x-15) &\mbox{ for $15\leq x \leq 20$}
                 \end{array} \right. $  & $1$  & $[0,20]$ & 0 \\
 \hline
 $f_{17}$   & $ \left\{ \begin{array}{ll}
		          80(2.5-x) &\mbox{ for $0\leq x <2.5$} \\
                  64(x-2.5) &\mbox{ for $2.5\leq x <5$} \\
                  64(7.5 - x) &\mbox{ for $5\leq x \leq 7.5$} \\
                  28(x-7.5) &\mbox{ for $7.5\leq x <12.5$} \\
                  28(17.5-x) &\mbox{ for $12.5\leq x <17.5$} \\
                  32(x-17.5) &\mbox{ for $17.5\leq x <22.5$} \\
                  32(27.5-x) &\mbox{ for $22.5\leq x <27.5$} \\
                  80(x-27.5) &\mbox{ for $27.5\leq x <30$}
                 \end{array} \right. $  & $1$  & $[0,30]$ & 0  \\
 \hline
 $f_{18}$   & $sin^6(5\pi x)$  & $1$  & $[0,1]_D$ & 0  \\
\hline
 $f_{19}$   & $exp[-2log(2).(\frac{x-0.1}{0.8})^2]. sin^6(5\pi x)$  & $1$  & $[0,1]_D$ & 0  \\
 \hline
 $f_{20}$   & $sin^6(5\pi (x^{3/4}-0.05))$  & $1$  & $[0,1]_D$ & 0  \\
 \hline
 $f_{21}$   & $\begin{matrix}exp[-2log(2).(\frac{x-0.08}{0.854})^2].\\sin^6(5\pi(x^{3/4}-0.05))\end{matrix}$  & $1$  & $[0,1]_D$ & 0  \\
\hline
 $f_{22}$   & $(x_1^2 + x_2 - 11)^2 + (x_1 + x_2^2 - 7)^2 $  & $2$  & $[-10,10]_D$ & 0  \\
 \hline
 $f_{23}$   & $ [(4 - 2.1x_1^2 + \frac{x_1^4}{3})x_1^2 + x_1x_2 + (-4+4x_2^2)x_2^2]$  & $2$  & $\begin{matrix}  [1.9,1.9]x \\ [-1.1,1.1] \end{matrix}$  & -1.03163  \\
 \hline
 $f_{24}$   & $sin(x_1)sin^2(\frac{x_1^2}{\pi })+sin(x_2)sin^2(\frac{2x_2^2}{\pi}) $  & $2$  & $[0,\pi]_D$ & 0  \\
 \hline
 $f_{25}$   & $0.26(x_1^2+x_2^2)-0.48x_1x_2$  & $2$  & $[-10,10]_D$ & 0  \\
 \hline
\end{longtable}

\begin{longtable}{|c|c|c|c|c|c|c|}
\caption{Evaluation of EVSOA on Benchmark Equations} \label{Tsol}\\
\hline
\textbf{Name} & \textbf{} & \textbf{Mean} & \textbf{SD} & \textbf{Best Solution} & \textbf{Worst Solution} & \textbf{Error}\\
\hline
\endfirsthead
\multicolumn{7}{c}%
{\tablename\ \thetable\ -- \textit{Continued from previous page}} \\
\hline
\textbf{Name} & \textbf{} & \textbf{Mean} & \textbf{SD} & \textbf{Best Solution} & \textbf{Worst Solution} & \textbf{Error}\\
\hline
\endhead
\hline \multicolumn{7}{r}{\textit{Continued on next page}} \\
\endfoot
\hline
\endlastfoot
 $f_1$ & EVO & 0.49523 & 	0.12854	 & 0.2103 & 	0.8719 & 	0.49523 \\
  \cline{2-7}
   & PSO & 0.9394 & 	0.2273 & 	0.01728 & 	1.23 & 	0.9394 \\
  \cline{2-7}
   & GA & 1.93 & 	1.121 & 	1.15787 & 	1.675 & 	1.93 \\
  \hline
  $f_2$ & EVO & 0.45852 & 	0.014751 & 	0.0141 & 	0.7996 & 	0.45852 \\
  \cline{2-7}
   & PSO & 0.3482 & 	0.5245 & 	0.124 & 	0.6412 & 	0.3482 \\
  \cline{2-7}
   & GA & 1.246 & 	1.4223 & 	0.97542 & 	1.9465 & 	1.246 \\
  \hline
  $f_3$ & EVO & 0.39028 & 	0.09812 & 	0.0451 & 	0.7992 & 	0.39028 \\
  \cline{2-7}
   & PSO & 0.6273 & 	0.2897 & 	.02773 & 	.8892 & 	0.6273 \\
  \cline{2-7}
   & GA & 1.8283 & 	0.9376 & 	0.36532 & 	2.1334 & 	1.8283 \\
  \hline
  $f_4$ & EVO & 0.23191	 & 0.00456	 & 0.0921	 & 0.5012 & 	0.23191 \\
  \cline{2-7}
   & PSO & 0.0876 & 	0.087 & 	0.0069	 & 0.1834	 & 0.0876 \\
  \cline{2-7}
   & GA & 0.6555 & 	0.7964	 & 0.0454	 & 1.3793	 & 0.6555 \\
  \hline
  $f_5$ & EVO & 0.10234 & 	0.0123 & 	0.0019 & 	0.3298	 & 0.10234 \\
  \cline{2-7}
   & PSO & 0.1368 & 	0.179 & 	0.007 & 	0.599 & 	0.1368 \\
  \cline{2-7}
   & GA & 0.186 &	0.938 & 	0.002 & 	0.4380 & 	0.186 \\
  \hline
  $f_6$ & EVO & -1 &	0	& -1 &	-1 &	0 \\
  \cline{2-7}
   & PSO &  -1 &	0	& -1 &	-1 &	0 \\
  \cline{2-7}
   & GA &  -1 &	0	& -1 &	-1 &	0 \\
  \hline
  $f_7$ & EVO & 0.414 & 	0.002 & 	0.4	 & 0.478 & 	0.0250 \\
  \cline{2-7}
   & PSO & 0.419 & 	0.003 & 	0.4 & 	0.443 & 	0.419 \\
  \cline{2-7}
   & GA & 0.451 & 	.0012 & 	0.4 & 	0.518 & 	0.451 \\
  \hline
  $f_8$ & EVO & 3.01777 & 	0.06985	 & 3.0008 & 	3.0193 & 	0.0059 \\
  \cline{2-7}
   & PSO & 3.1345 & 	0.0865 & 	3.0000 & 	3.1202 & 	0.045 \\
  \cline{2-7}
   & GA & 3.2976 & 	0.6757	 & 3.0000 & 	3.3201 & 	0.0992 \\
  \hline
  $f_9$ & EVO & 0.4021 & 	0.0092 & 	0.101 & 	0.8278 & 	0.4021 \\
  \cline{2-7}
   & PSO & 1.2039 & 	0.0203 & 	0.202 & 	1.02 & 	1.2039 \\
  \cline{2-7}
   & GA & 0.9281	 & 0.2823	 & 0.23 & 	1.0384 & 	0.9281 \\
  \hline
  $f_{10}$ & EVO & 0.005687 & 	0.08952 & 	0.00023 & 	0.0362	 & 0.005687 \\
  \cline{2-7}
   & PSO & 0.0175 & 	0.012	 & 0.00065 & 	0.056 & 	0.0175 \\
  \cline{2-7}
   & GA & 0.011	 & 0.97	 & 0.0081 & 	0.1875 & 	0.011 \\
  \hline
  $f_{11}$ & EVO & 0.007894	 & 0.001974	 & 2.45E-05	 & 0.014	 & 0.007894 \\
  \cline{2-7}
   & PSO & 0.00419	 & 0.0017	 & 0.0024	 & 0.093	 & 0.00419 \\
  \cline{2-7}
   & GA & 0.00751	 & 0.0340 & 	0.0001 & 	0.221 & 	0.00751 \\
  \hline
  $f_{12}$ & EVO & 0.2345 & 	0.1291 & 	0.01089	 & 0.9212 & 	0.2345 \\
  \cline{2-7}
   & PSO & 1.024	 & 0.0512	 & 0.000411	 & 1.6354	 & 1.024 \\
  \cline{2-7}
   & GA & 0.9756	 & 0.9842 & 	0.18753 & 	1.3544 & 	0.9756 \\
  \hline
  $f_{13}$ & EVO & -24771.1	 & 3.98730	 & -24776	 & -24765	 & 2.4e-4 \\
  \cline{2-7}
   & PSO & -24776	 & 0	 & -24776	 & -24776	 & 0 \\
  \cline{2-7}
   & GA & -24776 & 	0	 & -24776	 & -24776	 & 0 \\
  \hline
  $f_{14}$ & EVO & -1.9075	 & 0.003255	 & -1.9133	 & -1.9021	 & 0.0030 \\
  \cline{2-7}
   & PSO & -1.9058	 & 0.0012	 & -1.9133	 & -1.8292	 & 0.0039 \\
  \cline{2-7}
   & GA & -1.8974	 & 0.0028	 & -1.9133	 & -1.7096	 & 0.0083 \\
  \hline
  $f_{15}$ & EVO & 0.0002024	 & 0.000887	 & 0	 & 0.0029	 & .0002024 \\
  \cline{2-7}
   & PSO & 0.00057 & 	0.0003	 & 0	 & 0.0128	 & 0.00057 \\
  \cline{2-7}
   & GA & 0.0021	 & 0.00098	 & 0	 & 0.011	 & 0.0021 \\
  \hline
  $f_{16}$ & EVO & .000523 & 	0.000122 & 	0	 & 0.0082	 & 0.000523 \\
  \cline{2-7}
   & PSO & 0.00399 & 	0.02 & 	0	 & 0.0142 & 	0.00399 \\
  \cline{2-7}
   & GA & 0.00022	 & 0.001	 & 0	 & 0.0039 & 	0.00022 \\
  \hline
  $f_{17}$ & EVO & 0.001132	 & 0.000787	 & 0	 & 0.0039	 & 0.001132 \\
  \cline{2-7}
   & PSO & 0.07412	 & 0.0025 & 	0.0042 & 	0.4722 & 	0.07412 \\
  \cline{2-7}
   & GA & 0.01423	 & 0.0412	 & 0.00096	 & 0.1452	 & 0.01423 \\
  \hline
  $f_{18}$ & EVO & 0 &	0	& 0	& 0	& 0 \\
  \cline{2-7}
   & PSO &  0 &	0	& 0	& 0	& 0 \\
  \cline{2-7}
   & GA &  0 &	0	& 0	& 0	& 0 \\
  \hline
  $f_{19}$ & EVO & 0 &	0	& 0	& 0	& 0 \\
  \cline{2-7}
   & PSO & 0 &	0	& 0	& 0	& 0 \\
  \cline{2-7}
   & GA & 0 &	0	& 0	& 0	& 0 \\
  \hline
  $f_{20}$ & EVO & 0 &	0	& 0	& 0	& 0 \\
  \cline{2-7}
   & PSO & 0 &	0	& 0	& 0	& 0 \\
  \cline{2-7}
   & GA & 0 &	0	& 0	& 0	& 0 \\
  \hline
  $f_{21}$ & EVO & 0 &	0	& 0	& 0	& 0 \\
  \cline{2-7}
   & PSO & 0 &	0	& 0	& 0	& 0 \\
  \cline{2-7}
   & GA & 0 &	0	& 0	& 0	& 0 \\
  \hline
  $f_{22}$ & EVO & 0.0895 & 	0.0187	 & 0	 & 0.714	 & 0.0895 \\
  \cline{2-7}
   & PSO & 0.0438	 & 0.00156 & 	0	 & 0.112 & 	0.0438 \\
  \cline{2-7}
   & GA & 0.0841	 & 0.4753	 & 0 & 	0.9742	 & 0.0841 \\
  \hline
  $f_{23}$ & EVO & -1.0274	 & 0.01456	 & -1.03158	 & -1.02497	 & 0.0042 \\
  \cline{2-7}
   & PSO & -1.02942	 & 0.0389	 & -1.03163	 & -1.0192	 & 0.0021 \\
  \cline{2-7}
   & GA & -1.0212	 & 0.1736	 & -1.03158	 & -1.0142	 & 0.0198  \\
  \hline
  $f_{24}$ & EVO & 0 &	0	& 0	& 0	& 0 \\
  \cline{2-7}
   & PSO & 0 &	0	& 0	& 0	& 0 \\
  \cline{2-7}
   & GA & 0 &	0	& 0	& 0	& 0 \\
  \hline
  $f_{25}$ & EVO & 0.01582	 & 0.011987	 & 0	 & 0.0812	 & 0.01582 \\
  \cline{2-7}
   & PSO & 0.0554	 & 0.12551	 & 0	 & 0.18455 & 	0.0554 \\
  \cline{2-7}
   & GA & 0.018794 & 	0.05424	 & 0	 & 0.1541	 & 0.018794 \\
  \hline
\end{longtable}
\normalsize

%\subsection{Hybridization of GHOSA \& GA}

\section{Conclusions \& Future Works}
\label{sec-conclusion}
The work has been an extension for elaboration the GHOSA and its various operations both for the discrete and continuous optimization problems with the introduction of an enhanced and adaptive parameter variation factor (LBNIV) for only the continuous domain problems. The unique operations of the GHOSA naturally favour the discrete operations due to the natural behaviour of the algorithm for discrete event replacement in several schemes and varying the solutions and hence are very successful for most of the combinatorial optimization problems and other graph based challenges and optimization schemes. 
% Later the GHOSA has been extended to introduce the addition of opportunistic crossover of the Genetic Algorithm and thus produced an efficient hybrid structure. 
The result of application of the GHOSA has been compared with PSO, GA and ACO and has produced the potential of the algorithm for optimization in every sector of the problems. But yet a lot of performance analysis and convergence testing is required for the algorithm for other domains and other problem types like real time problems, dynamic and adaptive problems, etc. It is assumed that the algorithm can be applied to a lot of problems and analysis which require optimization. The main advantage of GHOSA is its adaptability for dimensionless (no proper dimension is there) problems like path planning and adaptive clustering (with unknown number of clusters). Also the GHOSA algorithm is compatible with both addition and removal of dimension(s) from the solution and this is the reason the algorithm will have a wide range of applicability in various domains of problems.

From the Table 1 to Table 5 and graphs of the result of various applications have revealed how the new meta-heuristics has delivered for optimization of the various dimensional benchmark datasets and its capability as discrete solution seeker and with the addition of the location based neighbour influenced variation scheme the algorithm has been capable to performed for the optimization of the continuous problems like numerical benchmark equations as well and can compete with traditional swarm intelligence algorithms. There awaits a lot of work on the new algorithm for the real world problems and its comparison with other algorithms.    
%\section*{Acknowledgements}

%\section*{Appendx}

%\end{document}

%\def\notesname{Note}
%
%\theendnotes

%\section*{Query}
%
%\tc{AQ1: AUTHOR PLEASE CITE FIGURE 8 IN TEXT.}

\end{document}